\documentclass[lettersize,journal]{IEEEtran}
\usepackage{amsmath,amsfonts}
\usepackage{algorithmic}
\usepackage{algorithm}
\usepackage{array}
\usepackage[caption=false,font=normalsize,labelfont=sf,textfont=sf]{subfig}
\usepackage{textcomp}
\usepackage{stfloats}
\usepackage{url}
\usepackage{verbatim}
\usepackage{graphicx}
\usepackage{booktabs}
\usepackage{tabularx}
\usepackage{makecell}
\usepackage{threeparttable}
\usepackage[numbers,sort]{natbib}
\usepackage{hyperref}

\begin{document}

\title{A Survey on Human Preference Learning for Large Language Models}

\author{Ruili Jiang, Kehai Chen, Xuefeng Bai, Zhixuan He, Juntao Li, 

Muyun Yang, Tiejun Zhao, Liqiang Nie, and Min Zhang
\thanks{Ruili Jiang, Muyun Yang, Tiejun Zhao are with the Faculty of Computing, Harbin Institute of Technology, Harbin, Heilongjiang, China (e-mail: ruilijiang@outlook.com; yangmuyun@hit.edu.cn; tjzhao@hit.edu.cn).}%
\thanks{Kehai Chen, Xuefeng Bai, Zhixuan He, Liqiang Nie, Min Zhang are with the School of Computer Science and Technology, Harbin Institute of Technology (Shenzhen), Shenzhen, Guangdong, China (e-mail: chenkehai@hit.edu.cn, baixuefeng@hit.edu.cn, hezhixuan1997@gmail.com, nieliqiang@hit.edu.cn, zhangmin2021@hit.edu.cn).}
\thanks{Juntao Li is with the Institute of Artificial Intelligence, Soochow University, Suzhou, Jiangsu, China (e-mail: ljt@suda.edu.cn).}%
}

\markboth{Journal of \LaTeX\ Class Files,~Vol.~14, No.~8, August~2021}%
{Shell \MakeLowercase{\textit{et al.}}: A Sample Article Using IEEEtran.cls for IEEE Journals}

\IEEEpubid{This work has been submitted to the IEEE for possible publication. Copyright may be transferred without notice, after which this version may no longer be accessible. }

\maketitle

\begin{abstract}
    The recent surge of versatile large language models (LLMs) largely depends on aligning increasingly capable foundation models with human intentions by preference learning, enhancing LLMs with excellent applicability and effectiveness in a wide range of contexts.
    Despite the numerous related studies conducted, a perspective on how human preferences are introduced into LLMs remains limited, which may prevent a deeper comprehension of the relationships between human preferences and LLMs as well as the realization of their limitations.
    In this survey, we review the progress in exploring human preference learning for LLMs from a preference-centered perspective, covering the sources and formats of preference feedback, the modeling and usage of preference signals, as well as the evaluation of the aligned LLMs. 
    We first categorize the human feedback according to data sources and formats.
    We then summarize techniques for human preferences modeling and compare the advantages and disadvantages of different schools of models.
    Moreover, we present various preference usage methods sorted by the objectives to utilize human preference signals.
    Finally, we summarize some prevailing approaches to evaluate LLMs in terms of alignment with human intentions and discuss our outlooks on the human intention alignment for LLMs.
\end{abstract}


\section{Introduction}
\IEEEPARstart{L}{arge} language models (LLMs) \cite{gpt3,anth-llm,ernie,chinch,glm,gpt4,palm2,llama2,qwen,jiang2024mixtral} have posed a groundbreaking impact on artificial intelligence (AI), transforming the opinions of people on the potential of AI systems for understanding and applying human languages. These neural network language models with large-scale parameters (mainly over 10 billion) are initially pre-trained on large corpora collected from a wide range of sources, a remarkable part of which is on the Internet \cite{llmsurv}. After pre-training by imitating how humans use natural languages in the text data, the foundation LLMs acquire strong and general language skills \cite{gpt3,zhou2023lima}.
On the other hand, foundation LLMs are observed to have difficulty in understanding or responding to diverse human instruction appropriately \cite{instgpt}, as the imitation process in pre-training does not enforce the foundation LLMs to follow the instructions as humans intended \cite{l2sum,instgpt}. 
Some toxic, biased, or factually incorrect content from the Internet left in the pre-training corpora would even lead to improper imitation of the foundation LLMs, resulting in undesirable generations \cite{realtox,weidinger2021ethical,kenton2021alignment,bommasani2022opportunities,dedup,gopher}. For practical applications in real life, the foundation LLMs must evolve to be more {\em aligned} with the intentions of humans, rather than {\em misaligned} imitation of potentially noisy behavior in the pre-training corpora.

\begin{figure*}[!t]
    \centering
    \includegraphics{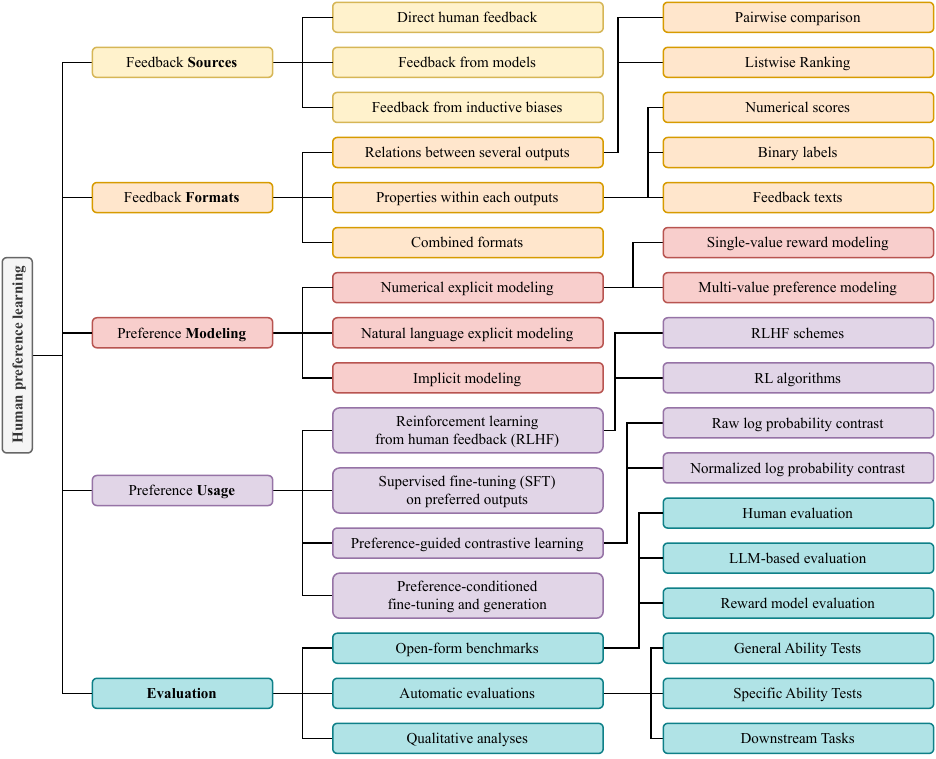}
    \caption{Overview of the aspects of human preference learning included in this survey.}
    \label{fig:tree}
\end{figure*}

\IEEEpubidadjcol
Human preference learning \cite{christianoDeepReinforcementLearning2017a} can effectively align LLMs with human intentions by optimizing LLMs according to feedback information on their outputs that reflects the preferences and thus specifies the intentions of humans \cite{aialign}. 
The effectiveness is validated by a recent surge of the evolved LLMs capable of generating proper responses to various human instructions \cite{instgpt,anth-llm,gpt4,llama2,qwen}. 
Currently, there are surveys on either a narrow approach for human preference learning or broad language model (LM) alignment. 
Surveys on human preference learning focus on reinforcement learning (RL), which may neither apply to LLMs nor contain insights related to non-RL preference learning approaches \cite{wirth2017survey,kaufmann2023survey}. 
Surveys on LM alignment \cite{shen2023large,aligning,kirk2023the,bridg}, as well as the alignment of general AI systems \cite{aialign} or big models beyond language \cite{wang2024essence}, mainly consider human preference learning as a tool to solve alignment problems. They lack a systematic review and discussion on preference learning, especially preference modeling methods, which are critical to capturing human intentions for LM alignment \cite{instgpt}.
To further explore effective preference learning approaches for better LLM alignment, we present a comprehensive review of human preference learning methods applicable to language models, examining LLM alignment methods from the perspective of preference learning. 
By analyzing a wide range of alignment approaches within the preference learning framework, we outline the holistic picture of introducing human preference into LLMs, enabling insights to be drawn from every aspect of human preference learning for various domains.

Specifically, the aspects we introduce human preference learning for LLMs, as shown in Fig. \ref{fig:tree}, include the {\em sources} and {\em formats} of preference feedback, the {\em modeling} of human preferences, the {\em usage} of preference signals, and the {\em evaluation} of human preference integrated LLMs:

\begin{itemize}
    \item {\em Feedback sources:} The quality and scale of preference feedback are of great importance for human preference learning, while the sources of feedback collection can heavily influence them. Recent human preference learning methods collect preference feedback from not only humans but also simulations of humans, exploring the balance between high-quality and large-scale.
    \item {\em Feedback formats:} The formats of preference feedback determine its information density and collection difficulty, thereby also impacting the quality and scale of preference feedback. The feedback formats adopted in works on human preference learning broadly include relative relations that are natural for preference expression but less informative, and absolute properties that are more informative about human preferences but harder to collect. The combinations of different formats can further increase the information density of preference feedback.
    \item {\em Preference modeling:} Preference modeling aims to obtain preference models from preference feedback, providing generalizable and directly usable human preference signals for aligning LLMs. Various preference modeling methods focus on obtaining preference models with numerical outputs. Some works also explore modeling preference methods with natural language outputs. Besides explicitly obtaining any preference model, another line of research implicitly models human preferences by directly using feedback data as preference signals to align LLMs with indirect preference modeling objectives or utilizing aligned LLMs to provide preference signals.
    \item {\em Preference usage:} Preference usage is the stage to adjust the foundation LLMs with the guidance of preference signals, aligning LLMs with human intentions. According to the specific objective of preference signal usage, recent methods can be divided into four main categories: \textit{reinforcement learning with human feedback (RLHF)} that maximizes the overall expected reward scores of LLM outputs; \textit{supervised fine-tuning (SFT) on preferred outputs} that maximizes the generation probabilities of the human-preferred output samples; \textit{preference-guided contrastive learning} that increases the generation probabilities of the more preferred outputs while decreasing the less preferred ones; and \textit{preference-conditioned fine-tuning and generation} that maximizes the generation probabilities of the outputs conditioned by corresponding preference signals.
    \item {\em Evaluation:} Finally, a comprehensive evaluation of the human-intention-following ability of LLMs is vital to verify the effectiveness of human preference learning. The prevailing evaluation protocols fall into three categories: open-form benchmarks that evaluate human preference for the responses of LLMs to diverse instructions without golden answers, automatic evaluations that evaluate LLMs with automatic metrics on sets of tasks with golden labels, and qualitative analyses that directly examine each response to some representative instructions.
\end{itemize}

Notably, the coverage of this survey includes research works on human preference learning that are not LLM-specific but can be applied to aligning LLMs, providing insights from fields such as classic reinforcement learning. We further summarize the key points of the recent advances in aligning LLMs with human intentions, and discuss the currently unsolved challenges and possible promising directions for future research, including pluralistic human preference learning, scalable oversight for aligning LLMs, language-agnostic LLM alignment, alignment with multi-modal complement, comprehensive assessment of LLM alignment progress, and empirically researching deceptive alignment. 
We hope this survey can help researchers discover the underlying mechanisms of how human preferences operate in LLM alignment, enlightening them on aligning LLMs and other AI systems with human intentions through the review of cutting-edge research works.

The organization of the rest of this survey is as follows. We begin with the background of this survey in Section \ref{sec:dev}, introducing the development timeline of human preference learning for LLMs. Then, we introduce aspects of human preference learning for LLMs from Section \ref{sec:sources} to Section \ref{sec:eval}, including feedback sources (Section \ref{sec:sources}), feedback formats (Section \ref{sec:formats}), preference modeling (Section \ref{sec:model}), preference usage (Section \ref{sec:hfutil}), and evaluation (Section \ref{sec:eval}). Last but not least, we conclude this paper with a summary of human preference learning and a discussion about our future outlooks in Section \ref{sec:conc}.

\begin{figure*}[!t]
    \centering
    \includegraphics{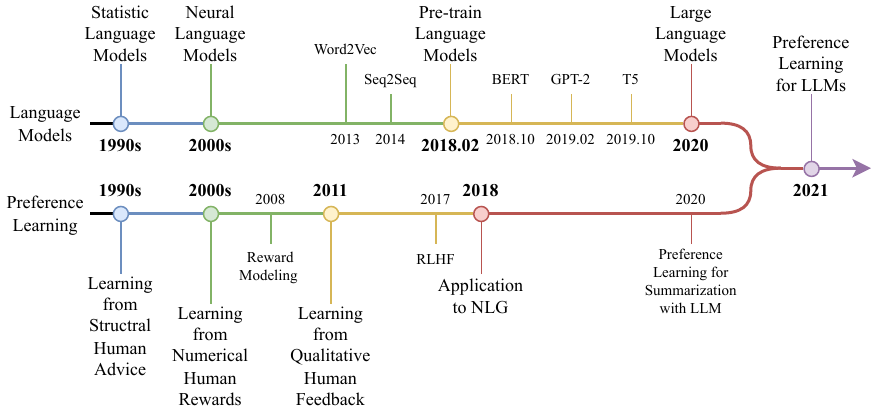}
    \caption{A timeline of the milestones in the development of human preference learning for LLMs.}
    \label{fig:timel}
\end{figure*}

\section{Background}
\label{sec:dev}

In this section, we provide the development timeline related to human preference learning for LLMs, including language modeling, preference learning, and their combination. The milestones in the timeline are depicted in Fig. \ref{fig:timel}.

\subsection{Language Models}

Language models (LMs) aim to learn the probability distribution of natural language based on the likelihood of generating a given text segment, beginning with the early form of {\em statistical language models (SLMs)} since the 1990s \cite{chu2024history,llmsurv,gao2004slm,rosenfeld2000twod}. Most SLMs are $n$-gram models \citet{bahl1983mlsr}, predicting the subsequent word tokens based on $n$ previous words as conditional contexts. In practice, these SLMs suffer from the curse of dimensionality since increasing $n$ leads to exponential growth of the parameters to estimate. 
Smoothing techniques were proposed to mitigate this, improving the SLMs \cite{zhai2004study,chen1999empirical,chen1996empirical}.

The {\em Neural Network Language Model} \cite{bengio2000neural,bengio2003neural} is a milestone in the development of LMs. It pioneered the integration of neural networks into LMs for learning distributed representations of words. 
Further, \citet{mikolov2013efficient} proposed the Continuous Bag-of-Words (CBOW) model and the Continuous Skip-gram model to learn the representation vectors of words, forming the basis of Word2Vec \cite{word2vec}. It significantly propelled the development and application of word vectors to many natural language processing (NLP) tasks \cite{nlpt,pennington2014glove}. 
The subsequent neural LMs following this trend applied recurrent neural networks (RNNs) and long short-term memory (LSTM) networks to obtain the word vectors, and further applied corresponding decoder networks for end-to-end sequence-to-sequence learning \cite{sutskever2014sequence,bahdanau2014neural}.

Another milestone in LM development is the ELMo 
representation \cite{Peters2018DeepCW}, pioneering the {\em pre-trained language models} by pre-training to learn contextually relevant word representations and then fine-tuning for specific downstream NLP tasks. 
After the emergence of the Transformer \cite{attisayn}, BERT \cite{devlin2018bert} was proposed, following the ``pre-training then fine-tuning'' scheme with a bidirectional Transformer encoder model. 
As its effectiveness is demonstrated in various NLP tasks, it established the ``pre-training then fine-tuning'' scheme as a paradigm for the subsequent development of LMs, followed by works including GPT-2 \cite{radford2019language} and T5 \cite{t5} with larger scales and different Transformer-based architectures. 

Following the trend of scaling up LMs, \citet{gpt3} proposed the GPT-3, the largest LM at the time. Besides enhanced performance on NLP tasks, the researchers found surprising extra features known as ``emergent abilities'' on GPT-3, which did not occur in smaller models \cite{wei2022emergent}. 
For example, it can imitate the demonstrations given in the input context to effectively complete various downstream tasks without further fine-tuning. Other research works on the increasingly large-scale pre-trained LMs also observed emergent abilities \cite{shanahan2023talking,cot,palm1}. To distinguish these models from the previous pre-trained LMs, the term ``{\em large language models (LLMs)}'' was coined, and GPT-3 was regarded as the milestone in the development of LMs that began the era of LLMs.

GPT-3 and most of the subsequent LLMs are pre-trained with the language modeling objective, predicting the next word token given any text segment as context \cite{gpt3,anth-llm,chinch,palm1,llama2,qwen}. It can be formalized as follows:
\begin{equation}
\label{equ:lm}
    \Phi_{LM}(\mathbf{x})=\sum\limits_{i = 1}^{|{\mathbf{x}}|} {\log P({x_i}|{{\mathbf{x}}_{ < i}})},
\end{equation}
where $\Phi_{LM}(\cdot)$ represent the language modeling objective function to maximize, $\mathbf{x}=\{x_1,x_2,\cdots,x_{|\mathbf{x}|}\}$ represent any tokenized sequence of text in pre-training corpora, $|\mathbf{x}|$ is the length of $\mathbf{x}$, and $P({x_i}|{{\mathbf{x}}_{ < i}})$ is the likelihood of the language model predicting $x_i$ given the preceding token sequence ${\mathbf{x}}_{ < i}$. Then, they generate output tokens $\{y_i\}$ autoregressively according to the input context $\mathbf{x}$ by sampling from $P({y_i}|\mathbf{x},{{\mathbf{y}}_{ < i}})$. The above shows that a pre-trained foundation LLM generates outputs according to the distribution learned from the pre-training corpora. On the other hand, the pre-training corpora do not necessarily include guidelines or examples on how to follow the intentions of humans \cite{l2sum,instgpt}, leading to the misalignment between the learned distribution and the desired output distribution. Some toxic, biased, or factually incorrect content in the pre-training corpora would further contaminate the learned distribution, causing the foundation LLMs to generate undesirable outputs \cite{realtox,weidinger2021ethical,kenton2021alignment,bommasani2022opportunities,dedup,gopher}.

\subsection{Preference Learning}
\label{subsec:prefl}
Preference learning is about automatically learning and predicting human preferences from feedback data with machine learning approaches to assist decision-making \cite{wirth2017survey,Fürnkranz2011}.
One of the early attempts at \textit{preference learning from structural human advice} can be traced back to the 1990s, in which an agent learned to integrate human feedback in programming languages into its action selection through specially designed neural networks \cite{maclin_creating_1996}. 

In the 2000s, some works started to focus on \textit{learning from numerical rewards provided by humans} \cite{bradknox-thesis}. For example, \citet{cobot0,cobot1,cobot2,isbell_cobot_2006} developed an agent on a social network that can learn from the statistics of rewards received from human users to adjust its behavior. \citet{tamer08,tamer09} proposed TAMER, a framework for explicitly modeling and predicting human rewards to perform complicated tasks in reinforcement learning \cite{tamer10,tamer12,tamer13}. Besides, reinforcement learning from human rewards was also applied to real-world robotics \cite{robot11p,robot11s,daniel_active_2015} and spoken dialogue systems \cite{sbirl,tdsha}.

One of the milestones in preference learning was the introduction of preference-based reinforcement learning (PbRL), focusing on \textit{learning from qualitative human feedback} that is easier to provide but harder to utilize than structural advice and numerical rewards \cite{ppl,pbpi01,furnkranz_preference-based_2012}. Many subsequent works followed the PbRL setting utilizing human preference comparisons as feedback, modeling human preference through ``learning to rank'' \cite{l2r} to guide the optimization of a policy \cite{ppl,april,progf} or form the policy itself \cite{pbpi01,furnkranz_preference-based_2012,pialpb13,Wirth_Fürnkranz_Neumann_2016}. \citet{wilson2021bayesian} alternatively formed a preference model from the policy so that the policy can be learned by preference modeling. \citet{wang-etal-2016-learning-language} explored learning from human feedback in natural language. 
Further, \citet{christianoDeepReinforcementLearning2017a} proposed reinforcement learning from human feedback (RLHF), paying more attention to learning a reward model with generalization ability from qualitative human feedback. 
\citet{ibarz2018reward} added a supervised pre-training step for the policy by imitation learning with expert demonstrations to warm-start RLHF, enhancing its learning efficiency. 

Another important step towards the combination of LMs and preference learning is \textit{the application of RLHF to natural language generation (NLG) tasks}, such as translation \cite{kreutzer-etal-2018-reliability}, review generation \cite{cho-etal-2019-towards}, summarization \cite{bohm-etal-2019-better,ziegler2020finetuning}, and stylistic continuation \cite{ziegler2020finetuning}. Some works also learned reward models with generalization ability for NLG training other than reinforcement learning \cite{hancock-etal-2019-learning,yi-etal-2019-towards,zhou_learning_2020}. \citet{l2sum} further incorporated LLMs for learning to summarize from human preference feedback. Then, \citet{askell2021general} made the first attempt to model task-agnostic human preference with LLMs, aiming to align LLMs with human values, including helpfulness, honesty, and harmlessness. \citet{instgpt} and \citet{anth-llm} subsequently experimented with applying RLHF to LLMs, opening the era of \textit{preference learning for LLMs}.

\section{Feedback Sources}
\label{sec:sources}

As a beginning of human preference learning, the sources of human feedback collection are critical since they are crucial to the quality and scale of preference feedback data representing human preferences. 
The feedback sources adopted by most works can be categorized into three types varying in quality and scalability: direct human feedback, feedback from LLMs, and feedback from inductive biases.

\subsection{Direct Human Deedback}

Intuitively, the most reliable sources of human preference feedback are humans. In practice, human labelers are typical sources of preference feedback, whether traditional preference learning \cite{maclin_creating_1996,cobot0,tamer08,ppl,christianoDeepReinforcementLearning2017a} or preference learning for LLMs \cite{instgpt,anth-llm,llama2,saferlhf,fghf}, whenever possible. Besides, many works on human preference learning for LLMs \cite{raft,dpo,steerlm} can also benefit from high-quality direct human feedback thanks to some open-accessible human-labeled feedback datasets, including HH-RLHF \cite{anth-llm}, OASST \cite{oasst}, and PKU-SafeRLHF \cite{beaver}. 
While human-labeled feedback mainly reflects the preferences of humans with high quality, the inter-individual differences within labelers and the misalignment between labelers and researchers may cause undesired problems, such as severe degeneration in feedback quality \cite{ziegler2020finetuning}. 
Considering that in most cases, labelers are crowd-workers or volunteers who are not so familiar with the research, research teams must keep connected with them, ensuring the alignment between preference annotations of labelers and the intentions of researchers \cite{instgpt,anth-llm}. 

\subsection{Feedback from Models}
\label{subsec:fbfmod}

While direct human feedback faithfully reflects the preferences of humans most of the time, some drawbacks limit its broad application to LLM alignment: i) hiring human labelers is too costly for most researchers, while open-accessible feedback datasets can not dynamically capture the human preferences of LLM outputs \cite{oasst,yang2023baichuan}; ii) scaling human feedback collection for the iterative update of reward model is time-consuming, which also hinders the up-to-date capture of human preferences to LLM outputs \cite{constai}. To alleviate these drawbacks, some researchers turn to exploring automatic simulation of human feedback with models, such as annotating preferences feedback with numerical reward models to be discussed in Subsection \ref{subsec:rewmod} \cite{gao23scaling,rrhf}.

On the other hand, LLMs become increasingly capable and aligned after learning from direct human feedback \cite{instgpt,gpt4,llama2}. Therefore, an emerging research trend is eliciting preferences from LLMs to simulate human preference feedback automatically. Constitutional AI \cite{constai} makes the first attempt to gather preference feedback regarding harmlessness, leveraging a foundation LLM prompted with human-written principles of harmlessness. SALMON \cite{salmon} further extends this idea to feedback collection regarding multiple aspects of human preferences, such as conciseness, honesty, and multilingualism. Many subsequent works \cite{farm,wizmath,ultrafb,guo2024beyond,rlmec} make use of powerful commercial LLMs (GPT-4 \cite{gpt4}, etc.) as simulated human feedback sources. \citet{rlaif} further verify that although more likely to be affected by hallucination or issues in coherence and grammar, the feedback gathered from LLMs can achieve or even surpass the human level in summarization and dialogue tasks. These works demonstrate the feasibility of feedback from models, especially LLMs, as an economical and efficient feedback source with minor quality issues.

\subsection{Feedback from Inductive Biases}

Inductive biases are sets of preferences, priors, or assumptions humans may exploit in the cognitive process \cite{indbias}. If simple rules can be extracted from the inductive biases of humans, they can also be regarded as simulated sources of human preference feedback. 
A typical application of this idea in preference learning for LLM is the preference model pre-training dataset collected in \cite{anth-llm}. Utilizing data dumps of online Q\&A communities, preference pairs are constructed based on the inductive bias that ``answers with larger user upvote numbers are more preferred.'' The SHP dataset \cite{shp} is collected in a similar way, considering both upvotes and downvotes with more careful data curation.

Another example is ALMoST \cite{almost}, ranking preferences of responses from LLMs with different response-generating conditions based on a set of rules: i) LLMs with more parameters generate better responses than ones with fewer parameters; ii) LLMs prompted with more in-context instruction following demonstrations generate better responses than ones with fewer in-context demonstrations; iii) LLMs prompted with in-context demonstrations of higher quality generate better responses than ones prompted with demonstrations of lower quality. 
Although the collected feedback from prior knowledge is noisy and requires heuristic filtering, it simulates human preference feedback as expected, without any additional cost for commercial LLMs or any inference time required by LLMs \cite{almost}. 
RLCD \cite{yang2024rlcd} construct preference comparison feedback based on a similar rule to rule ii) in ALMoST, asserting the responses generated by possibly prompted LLMs are always well-behaved and vice versa. OpenChat \cite{openchat} further asserts that humans prefer responses generated by GPT-4, which is more capable and aligned \cite{gpt4}, to those by GPT-3.5. According to this assertion, higher preference scores are assigned to GPT-4-generated responses than GPT-3.5-generated ones. 

Besides, APO \cite{apo} and SPIN \cite{chen2024selfplay} adopt the inductive bias similar to generative adversarial networks (GANs) \cite{gan}. Specifically, they always prefer reference responses annotated by humans (or simulated references from commercial LLMs) to LLM-generated ones, closing the gap between LLM-generated and human-preferred responses.

\section{Feedback Formats}
\label{sec:formats}

The feedback formats concretely represent human preferences for the outputs of models to be collected from feedback sources, determining the information density and collection difficulty of human preference feedback. 
Based on whether the format is relative or not, the feedback formats adopted by most works on preference learning include relations between several responses and properties within each response. Besides, the combined formats of these categories are also feasible.

\subsection{Relations Between Several Outputs}

Relations between several responses are the prevailing feedback formats adopted by recent works on preference learning methods. 
As discussed in Subsection \ref{subsec:prefl}, these qualitative preference feedback formats are easy to collect, while their information density is relatively low, making them harder to utilize \cite{ppl,pbpi01,furnkranz_preference-based_2012}. According to the number of related outputs in each feedback item, relations between several responses can be divided into pairwise comparison and listwise ranking.

\subsubsection{Pairwise comparison}

Pairwise comparison specifies the ``one is better than the other'' relations between two outputs as feedback. 
It is widely adopted by preference learning methods, including both traditional preference learning \cite{ppl,progf,christianoDeepReinforcementLearning2017a} and preference learning for LLMs \cite{anth-llm,dpo,tara,apo}. 
\citet{webgpt} considers the case of tied comparisons, where ``one is as good as the other''. A special case of pairwise comparison in preference learning for LLMs is response editing, including improvement editing and adversarial editing. Improvement editing focuses on improving the LLM-generated responses, correcting the errors by editing the original responses with the fewest possible changes \cite{guo2024beyond,rlmec}. Adversarial editing focuses on generating adversarial negative responses from reference positive responses, applying limited perturbations to reference responses to distort their original meanings \cite{sinha2024break}. The edited responses can, therefore, naturally construct pairwise comparisons with the original responses to be used for targeted LLM alignment in reliability and safety.

\subsubsection{Listwise ranking}

Listwise ranking is the intuitive extension of pairwise comparison when more than two outputs are available. It is also adopted by a great number of works on both traditional preference learning \cite{pbpi01,furnkranz_preference-based_2012,brown2019extrapolating} and preference learning for LLMs \cite{rrhf,song_preference_2024,wizmath}. Among those, a typical example is InstructGPT \cite{instgpt}, a series of aligned LLMs whose preference alignment method is applied to the powerful ChatGPT\footnote{ \url{https://openai.com/blog/chatgpt/}} with dialogue-formatted instruction data. 
Compared to pairwise comparison, listwise ranking is more informative and similarly easy for collection \cite{instgpt}. 
On the other hand, similar to pairwise comparison, some crucial aspects of human preference information are still missing, including the degree or probability of preference \cite{llama2,rlaif}, the reasons behind the preference \cite{ultrafb}, etc.

\subsection{Properties Within Each Output}

In complement to the relations between several outputs, properties within each output provide more details about human preferences, revealing more critical aspects of human preference information \cite{bridg}. 
Specifically, common properties can be roughly classified into the following three categories: numerical scores, binary labels, and feedback texts.

\subsubsection{Numerical scores}

Numerical scores are widely used to describe the degree of human preferences, also known as rewards in traditional preference learning discussed in Subsection \ref{subsec:prefl}. In the context of preference learning for LLMs, numerical scores can be either continuous \cite{rlaif,openchat,guo2024beyond} or discrete \cite{ultrafb,oasst}. The continuous scores are usually manually assigned constants \cite{openchat,guo2024beyond} or obtained from models. For example, \citet{rlaif} utilize prediction probabilities of specific output tokens of an LLM as preference scores. On the other hand, discrete scores are usually annotated in similar forms to Likert scales \cite{likert_technique_1932}, including sets of descriptions about different degrees of human preference and their corresponding scores. \citet{oasst} annotate multiple discrete preference scores describing various aspects of human preferences for each response with human labelers. \citet{ultrafb} utilize a commercial LLM to annotate discrete preference scores at scale.

\subsubsection{Binary labels}

Binary labels are used to denote whether certain aspects of human preferences are satisfied or not \cite{saferlhf,fghf,bayes,ulma,ethayarajh2024kto}. They are on the borderline of qualitative feedback and numerical feedback as the exceptional cases of discrete reward scores, possible to be modeled with the classification probabilities of preference discriminators as in GANs \cite{gan}, differently from numerical rewards \cite{fghf,ulma}. As for the related preference modeling approaches, we will discuss them in detail in Section \ref{subsec:rewmod}.

\subsubsection{Feedback texts}

Feedback texts are qualitative feedback formats used to criticize or refine the responses based on human preferences in natural language \cite{ultrafb,tara}. For example, \citet{wang2023shepherd} curated a feedback text dataset from online Q\&A societies with human annotations that judge the weaknesses of the provided answers and offer suggestions to improve them. Similarly, \citet{ultrafb} collect critique texts for LLM-generated responses. Besides, \citet{tara} invoke a tool-augmented validation process for each answer, regarding its text record as the feedback that indicates the correctness of the answer.

\subsection{Combined Formats}

Combining different feedback formats is a simple approach to complementing the shortcomings of each format. 
For example, \citet{ultrafb} curate a preference feedback dataset combining numerical preference scores and the corresponding feedback texts that provide supporting reasons for the scores. 
\citet{rlaif} complement pairwise comparisons with their preference probabilities. \citet{llama2} annotate discrete preference levels for each pairwise comparison, which can also be regarded as numerical scores. \citet{tara} provide the reasoning text records of the tool-augmented validation process for both positive and negative answers in each pairwise comparison. \citet{saferlhf} assign binary labels of whether each LLM response is harmless in the absolute sense for pairwise comparisons besides their relative relations of both helpfulness and harmlessness. 
\citet{guo2024beyond} assign constant reward scores to edited tokens in the pairwise comparisons constructed by response editing, providing token-wise supervision for preference learning. \citet{fghf} further combine pairwise comparisons regarding information completeness of the full responses and binary labels regarding relevance and factuality in segments of responses, forming the multi-aspect fine-grained human preference feedback. 

\section{Preference Modeling}
\label{sec:model}
Preference modeling aims to obtain preference models from preference feedback data. 
While it is viable to directly apply the collected preference feedback as usable preference signals \cite{dpo,zheng2023improving,openchat}, a typical practice is to learn preference models with generalization ability as proxies of human preferences, providing signals for preference usage at scale \cite{gao23scaling,farm}. 
Recent works model human preferences in three major categories: numerical explicit modeling, non-numerical explicit modeling, and implicit modeling.

\subsection{Numerical Explicit Modeling}
\label{subsec:rewmod}

Numerical explicit modeling is typical in preference modeling, focusing on obtaining preference models whose outputs are numerical preference signals for preference usage. 
We classify numerical explicit modeling approaches into single-value and multi-value reward modeling depending on the output information density of the numerical preference models.

\subsubsection{Single-value reward modeling}

Numerical preference models with scalar outputs are also regarded as reward models, and learning reward models are known as reward modeling \cite{instgpt}. 
A reward model can be formalized as a reward score function $h(\mathbf{x}, \mathbf{y})$ fitted by a neural network, where $\mathbf{x}$ denotes the input condition (such as human instruction, if applicable) and $\mathbf{y}$ denotes the model output. In the context of preference learning for LLMs, reward models commonly have the same or similar Transformer backbone as LLMs, whose hidden output of the last token is mapped to a scalar with a linear layer as the output reward score \cite{instgpt,anth-llm,llama2}. 
In the following, we introduce some typical single-value reward modeling approaches that provide one reward score for each item in the preference feedback data, categorized according to the data formats used in reward modeling:

\begin{itemize}

    \item {\em Comparisons or rankings:} A typical case of single-value reward modeling with preference feedback in pairwise comparisons is given in \cite{christianoDeepReinforcementLearning2017a}. Human preferences are modeled by maximizing the log probability that the reward score of the preferred output $\mathbf{y}^w$ with input condition $\mathbf{x}$ (if any) is higher than the other output $\mathbf{y}^l$, formalized as follows:
    \begin{equation}
    \label{equ:cpref}
        \log P({\mathbf{y}^w} \succ {\mathbf{y}^l}|\mathbf{x}) = \log \left( {\sigma \left( {h(\mathbf{x},{\mathbf{y}^w}) - h(\mathbf{x},{\mathbf{y}^l})} \right)} \right),
    \end{equation}
    where $P({\mathbf{y}^w} \succ {\mathbf{y}^l}|\mathbf{x})$ represents the probability that $\mathbf{y}^w$ is preferred to $\mathbf{y}^l$ related to reward score given the input $\mathbf{x}$, and $\sigma(\cdot)$ represents the sigmoid function. Many subsequent works directly follow this approach \cite{apo,l2sum,anth-llm,fghf}. InstructGPT \cite{instgpt} further extends it for the ranking format by exhausting all pairwise combinations as comparisons in ranking feedback data, followed by ALMoST \cite{almost}.  
    \citet{webgpt} further consider the golden preference distributions of the tied comparisons uniform, averaging the log probabilities that each response is preferred. 

    \item {\em Comparisons or rankings with numerical scores:} \citet{rlaif} largely follow the reward modeling approach shown in Equation \eqref{equ:cpref}, while regarding the collected preference probabilities in the feedback as the golden preference distributions to weight the summation of the log probabilities. 
    The reward models of Llama-2-Chat \cite{llama2} and UltraRM \cite{ultrafb} instead introduce discrete scores representing the preference levels of the comparisons as a margin term into Equation \eqref{equ:cpref}. This modifies Equation \eqref{equ:cpref} to the following:
    \begin{align}
    \label{equ:cpmod}
        \log P({\mathbf{y}^w} \succ {\mathbf{y}^l}|\mathbf{x}) &= \log \left( {\sigma \left( {h(\mathbf{x},{\mathbf{y}^w}) - h(\mathbf{x},{\mathbf{y}^l})} \right.} \right. \nonumber \\
        & \left. {\left. {- \alpha m(\mathbf{y}^w,\mathbf{y}^l,\mathbf{x})} \right)} \right),
    \end{align}
    where $m(\mathbf{y}^w,\mathbf{y}^l,\mathbf{x})$ denotes the score representing the preference level of the comparison comprising outputs $\mathbf{y}^w,\mathbf{y}^l$ and input $\mathbf{x}$, and $\alpha$ is the scaling coefficient. 
    The margin term enforces a more significant difference in reward scores for responses with a larger quality gap.
    
    \item {\em Comparisons with binary labels:} The safety reward model used in Safe RLHF \cite{saferlhf} incorporates an extra binary classification objective according to the binary safety labels alongside the typical reward modeling objective shown in Equation \eqref{equ:cpref}. 
    This additional objective enforces the reward scores of responses with different binary safety labels to be divided by the reference value of zero.

    \item {\em Comparisons with additional texts:} As most reward models share a backbone architecture similar to LLMs, enhancing them by enriching the input context with additional texts is possible, akin to LLM prompting \cite{munos2023nash,salmon,tara}. For example, \citet{munos2023nash} prepend a concise description of the preference rating task. SALMON \cite{salmon} incorporates human-written principles about aspects of human preferences into the context, enforcing the modeling of how the response follows these principles. \citet{tara} include the reasoning texts recording the tool-augmented validation processes of each answer as the additional context, enforcing the reward model to learn to score according to validations.

    \item {\em Other input formats:} A few works also try to obtain single-value numerical reward models given the preference feedback data without any comparison- or ranking-based format. For example, \citet{fghf} model the rewards in relevance and factuality as binary classification probabilities given feedback data formatted in binary labels. \citet{prm} use a similar classification-based modeling approach for each segment in outputs with a three-class label (including positive, negative, and neutral that is regarded as either positive or negative in test time).
    \citet{bayes} model the reward of any given response to the instruction as the posterior classification probability of human acceptance. To this end, a KL-diversity-based objective is derived using the variational Bayesian technique with the human-labeled acceptance probability values as prior probabilities.
    
\end{itemize}

\subsubsection{Multi-value preference modeling}

Although single-value reward modeling already works well in the works discussed above, there are cases where single-value reward modeling is not capable enough. For example, a single-value reward model may overfit to feedback data, struggle to cover multiple aspects of human preferences, or fail to provide proper preference signals for fine-grained segments of the output sequences \cite{rmensemb,anth-llm,constai,prm,fghf}. 
In the following, we discuss some works on introducing multi-value preference modeling to mitigate the issues mentioned above or explore new possibilities in preference modeling:

\begin{itemize}

    \item {\em Mitigating overfitting:} \citet{christianoDeepReinforcementLearning2017a} train different reward models on random subsets of the preference feedback dataset and average the separately normalized rewards from each reward model as the final reward score. \citet{rmensemb} further combine different reward models on the same feedback dataset with different random seeds, selecting the lowest reward score or adding a weighted term of negative variance to reduce overfitting on feedback data.

    \item {\em Modeling multi-aspect human preferences:} Constitutional AI \cite{constai} and Safe RLHF \cite{saferlhf} use two separate reward models to model human preferences in both helpfulness and harmlessness, effectively alleviating the trade-off between the two objectives discovered in \cite{anth-llm}. \citet{fghf} separately learn three reward models regarding relevance, factuality, and information completeness, respectively. \citet{rwsoup} combine reward models concerning multiple aspects of human preferences by weighted sums of the parameters of reward models, where the weights sum up to 1. The combined model with the highest average reward on validation samples is selected as the final reward model. SteerLM \cite{steerlm} instead directly models the discrete scores for multiple attributes concerning human preferences in text format. As a result, an ``attribute prediction model'' is obtained to simultaneously predict the scores of all the corresponding attributes similarly to the auto-regressive text generation of LLMs.
    
    \item {\em Providing fine-grained rewards:} \citet{prm} learn a process-supervised reward model that learns to predict the probability of correctness for each step in reasoning given the fine-grained binary labels of the correctness in each math reasoning step as preference feedback. 
    \cite{fghf} train reward models providing segment-wise rewards in relevance and factuality. \citet{rlmec} further train a generative reward model by imitating the response improvement editing process, using the generative editing probability of each token as the token-wise reward.

    \item {\em Explorations in vector-based preference modeling:} Besides scalar rewards, another possibility of preference signal is vector representation. 
    For example, \citet{rahf} explore human preference modeling with the activation patterns of LLMs in the form of vector representations. They extract activation vectors from LLMs fine-tuned on the responses either preferred or not preferred by humans as the ``preference model'' in the representation space of LLMs. \citet{frans2024unsupervised} further explore the possibility to encode unknown reward functions into representation vectors with corresponding data samples, guiding reinforcement training on diverse tasks.
    
\end{itemize}

\subsection{Natural Language Explicit Modeling}

Apart from numerical preference models, preference models providing natural language preference signals are also explored. For example, \cite{wang2023shepherd} and \cite{ultrafb} train ``critic models'' that model human preference by learning from critique texts presented in feedback data, enabling it to judge the responses and offer suggestions for correcting flaws and improving quality. The critic model provides judgments and suggestions in texts for human preference usage methods that make use of natural language feedback \cite{richardson-heck-2023-syndicom,xu2023reasons}. As mentioned above, the learning of tool-augmented validation ability of the reward model in \cite{tara} can also be regarded as preference modeling in the format of natural language feedback. \citet{akyurek-etal-2023-rl4f} further train a critique generation model by reinforcement learning, utilizing the similarity metrics between human-preferred responses and responses refined with generated critiques as rewards to optimize.

\subsection{Implicit Modeling}

Some approaches bypass the preference modeling process by directly using instructed LLMs for scalable and generalizable human preference signals. 
As discussed in Subsection \ref{subsec:fbfmod}, LLMs instructed with appropriate prompts are capable of providing human preference information. What differs from the above is that instructed LLMs as preference models aim to directly provide generalizable preference signals at scale for human preference usage, rather than supplying preference feedback data for preference modeling. In \cite{rlaif}, both schemes are verified, namely Distilled RLAIF (LLM feedback for preference modeling) and Direct RLAIF (instructed LLM for preference usage), respectively. \citet{tuna} instruct commercial LLMs to obtain preference ranking feedback, either an LLM for response generation with logits available for ranking or an LLM prompted directly for response evaluation. \citet{humpback} further utilize an LLM prompted with the quality evaluation instruction to obtain reward scores of the given instruction following data, which are then used to filter the data for the fine-tuning of the LLM itself.

\begin{figure*}[t]
    \centering
    \includegraphics[width=\textwidth]{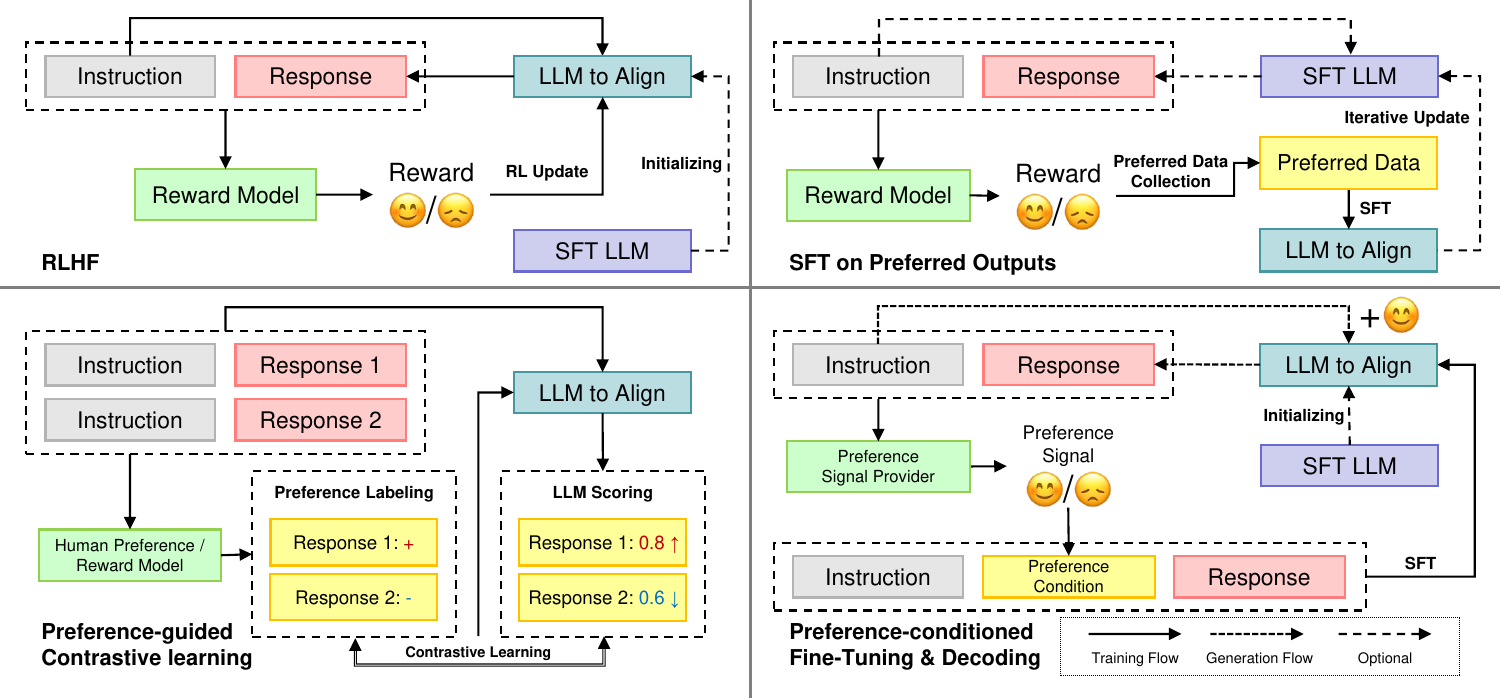}
    \caption{Illustrations of different preference usage paradigms, exemplified by LLM alignment.}
    \label{fig:util}
\end{figure*}

\section{Preference Usage}
\label{sec:hfutil}

Preference usage is the key stage to align the foundation LLMs with human intentions guided by collected or modeled preference signals. 
According to the specific objective of preference signal utilization, recent methods of human preference usage fall into four main categories: reinforcement learning from human feedback, supervised fine-tuning on preferred outputs, preference-guided contrastive learning, and preference-conditioned fine-tuning and generation. Fig. \ref{fig:util} illustrates the paradigms of each category.

\subsection{Reinforcement Learning from Human Feedback} 

Reinforcement learning from human feedback (RLHF) learns human-preferred models by reinforcement learning (RL) algorithms using numerical reward models trained from human preference feedback as preference signals \cite{instgpt,anth-llm,llama2,yang2024rlcd,rlmec}. 
Concretely, RL algorithms maximize the expected rewards of possible actions of the generative policy (e.g. outputs of an LLM) by iterating output sampling, reward evaluation, and model parameter updating with the Monte Carlo technique \cite{reinforce}, as demonstrated in the top-left of Fig. \ref{fig:util}. 
Apart from the general idea of RLHF discussed above, two modifications are commonly applied for improving effectiveness and efficiency \cite{instgpt,llama2}: i) before RL, supervised fine-tuning (SFT) on high-quality input-output demonstration pairs is usually conducted for the foundation model as initialization, forming the well-known three-staged RLHF scheme, i.e., SFT, reward modeling (discussed in Subsection \ref{subsec:rewmod}), and RL; ii) during RL, a KL divergence term between the RL policy and a reference model (in most cases, the SFT model) is appended as a regularizer to alleviate overfitting to the reward model.
To sum up, the formalization of the typical RLHF scheme is shown as follows:
\begin{equation}
    \begin{aligned}
        &\mathop {\max }\limits_\theta  \mathop \mathbb{E}\limits_{{\mathbf{x}} \sim D,{\mathbf{y}} \sim {\pi _\theta }({\mathbf{y}}|{\mathbf{x}})} h({\mathbf{x}},{\mathbf{y}}) - \beta \operatorname{KL} [{\pi _\theta }({\mathbf{y}}|{\mathbf{x}})||{\pi _{ref}}({\mathbf{y}}|{\mathbf{x}})] \\ 
        = &\mathop {\max }\limits_\theta  \sum\limits_{j = 1}^{|D_{\pi _\theta }|} {\left[ {h({{\mathbf{x}}^j},{{\mathbf{y}}^j}) - \beta \log \left( {{\pi _\theta }({{\mathbf{y}}^j}|{{\mathbf{x}}^j})/{\pi _{ref}}({{\mathbf{y}}^j}|{{\mathbf{x}}^j})} \right)} \right]},  \\ 
    \end{aligned}
\end{equation}
where $h(\mathbf{x}, \mathbf{y})$ denotes the output of the trained reward model given input $\mathbf{x}$ (if applicable) and output $\mathbf{y}$; $D$ denotes the dataset of inputs; ${\pi _\theta }$ denotes the output distribution of the policy model and reference model, respectively; $\beta$ is the coefficient of the KL divergence term, controlling the strength of regularization; and $D_{\pi _\theta }$ denotes the dataset of the inputs $\mathbf{x}^j$ and sampled outputs $\mathbf{y}^j$ from policy ${\pi _\theta }$.

Recent advances in RLHF approaches for preference usage mainly include improvements in RLHF schemes and RL algorithms. We will present each of them below.

\subsubsection{RLHF schemes}

One of the improving directions of the RLHF paradigm discussed above is to combine multiple reward scores related to different aspects of human preferences. For example, Safe RLHF \cite{saferlhf} combines the RLHF objectives of maximizing the helpfulness reward and restricting the safety cost (negative reward) with dynamic adjustment by applying the Lagrangian method \citet{lagrange}. \citet{crlhf} further combines multiple rewards given threshold constraints using the Lagrangian method to alleviate overfitting to all these rewards. \citet{wizmath} make use of the rewards modeling the quality of the LLM-augmented instructions as the weights for step-wise reward optimization to increase the correctness of the responses to the math reasoning instructions. Besides multi-reward RLHF, TS-LLM \cite{feng2023alphazero} introduces an AlphaZero-style tree search algorithm to guide the generation and RL-based fine-tuning of LLMs. NLHF \cite{munos2023nash} formalizes RLHF into a two-player game, where the policy model and a mixture model of the policy and the reference model compete and evolve together with RL until achieving a Nash equilibrium.

\subsubsection{RL algorithms}

In terms of RL algorithms, proximal policy optimization (PPO) is widely applied as it can significantly enhance the extent of alignment with human intentions for policy models \cite{schulman2017proximal}. 
On the other hand, the limitations of PPO are also prominent: i) it is highly dependent on hyperparameter tuning, which increases the risk of training instability; ii) the introduction of a learnable value model in its actor-critic scheme adds a massive burden on the computational resources required for the training process, especially VRAM of GPUs. To alleviate these drawbacks, studies on either improving PPO or substituting PPO are conducted.

\begin{itemize}

    \item {\em Improving PPO:} To stabilize PPO, P3O \cite{wu2023pairwise} introduces the pairwise difference of reward scores into the policy gradient computation, eliminating the potential variance in gradients caused by biased value estimation. 
    \citet{zheng2023improving} group the training data with a self-supervised model, strengthening the KL regularization of the data groups easy to optimize and encouraging the reward optimization of hard ones. This strategy strikes the balance between RL exploration and training stability. 
    CPPO \cite{anonymous2024cppo} preserves the knowledge learned by the policy model by incorporating a dynamically weighted knowledge reservation term into the PPO objective for stability when updating the policy model with new knowledge. 
    To lighten the burden of VRAM requirement for PPO, \citet{santacroce2023efficient} share the parameters of the backbone LLM with the policy model, the reference model, the reward model, and the value model, updating only the learnable parameters of lightweight LoRA modules for the policy and value model.
    
    \item {\em Substituting PPO:} Instead of refining PPO, ReMax \cite{li2023remax} removes the non-essential components in PPO under the background of auto-regressive language generation, reverting to a lightweight form akin to the REINFORCE algorithm \cite{reinforce}. It further incorporates a subtractive reward baseline for stability in training. \citet{rlaif} and \citet{ahmadian2024basics} also apply REINFORCE style RL algorithms to RLHF for LLMs.
    
\end{itemize}

\subsection{Supervised Fine-tuning on Preferred Outputs} 

In addition to RL, some SFT models trained on high-quality demonstration data also achieve remarkable performance \cite{wang2022self,salign,zhou2023lima,chen2024alpagasus}.  
An intuitive adaptation of SFT to the preference usage setting is to SFT on the dataset constructed from human-preferred outputs, as demonstrated in the top-right of Fig. \ref{fig:util}. This paradigm can be formalized as maximizing the generation probability of the preferred outputs, as shown below:
\begin{equation}
    \mathop {\max }\limits_\theta  \sum\limits_{j = 1}^{|{D_{pref}}|} {\log {\pi _\theta }({{\mathbf{y}}^j}|{{\mathbf{x}}^j})},
\end{equation}
where $D_{pref}$ denotes the dataset of the inputs $\mathbf{x}^j$ and human-preferred outputs $\mathbf{y}^j$.
As an example, RAFT \cite{raft} aligns LLMs by iterating the generation of responses to instructions, evaluating reward scores through a reward model, filtering by selecting the response with the highest reward for each instruction, and SFT on the filtered samples preferred by the reward model. The training of Llama-2-Chat \cite{llama2} includes a similar training stage as the warm start of the PPO training stage. 
SuperHF \cite{mukobi2023superhf} further includes an additional KL divergence term for regularization, similar to RLHF. 
\citet{humpback} explore the generation of instructions given massive unlabeled corpora as known responses, forming synthetic instruction-response data which are then iteratively filtered by an LLM for the SFT of itself. 
Besides, IFT \cite{hua2024intuitive} performs SFT on ``tokens'' mixing both LLM-generated tokens and supervised label tokens that are preferred by humans in the input embedding layer, providing a smooth learning process for LLMs to adapt to the human-preferred output distribution.

\subsection{Preference-guided Contrastive Learning}

Although SFT on preferred output fosters models to generate what humans prefer, merely imitating preferred outputs may not prevent the models from generating what humans do not prefer \cite{llama2,liu2024chain}. 
An alternative preference usage paradigm using the less preferred responses is preference-guided contrastive learning, as demonstrated in the bottom left of Fig. \ref{fig:util}. 
In addition to SFT on preferred output, preference-guided contrastive learning decreases the generation probabilities of the less preferred outputs while still increasing the more preferred ones, formalized as follows:
\begin{equation}
    \mathop {\max }\limits_\theta  \sum\limits_{j = 1}^{|{D_{con}}|} {g\Big( {{\beta _1}f\left( {{\pi _\theta }({{\mathbf{y}}^{j, + }}|{{\mathbf{x}}^j})} \right) - \sum\limits_{{{\mathbf{y}}^{j, - }}} {{\beta _2}f\left( {{\pi _\theta }({{\mathbf{y}}^{j, - }}|{{\mathbf{x}}^j})} \right)} } \Big)} ,
\end{equation}
where $D_{con}$ denotes the labeled preference dataset with inputs $\mathbf{x}^j$, positive outputs $\mathbf{y}^{j,+}$ (more preferred, usually one per sample), and negative outputs $\mathbf{y}^{j,-}$ (less preferred, can be more than one per sample); $f(\cdot)$ and $g(\cdot)$ are nondecreasing functions; $\beta_1$ and $\beta_2$ are the weights for positives and negatives, respectively. Depending on the concrete form of $f(\cdot)$, most preference-guided contrastive learning approaches can be divided into two types: raw log probability contrast and normalized log probability contrast.

\subsubsection{Raw Log Probability Contrast}

Raw log probability contrast is the case where $f(u)=log(u)$ for a scalar $u$.
The simplest example is unlikelihood training \cite{li-etal-2020-dont,Welleck2020Neural,secth,guo2024beyond}, directly decreasing the log probabilities of the negatives while increasing the log probabilities of the positives, i.e., $g(\cdot)$ is an identity function. Unlikelihood training can also be regarded as a direct extension of SFT on preferred output. 
Besides, any loss functions applicable to contrastive learning can be applied as the negative of $g(\cdot)$, including hinge loss \cite{rrhf,zhao2023slichf,bayes,tuna,liu2024training}, InfoNCE (similar to reward modeling objective in Equation \eqref{equ:cpref})\cite{wang2023making,song_preference_2024,jiang2024preference}, or modified reward modeling objective in Equation \eqref{equ:cpmod} \cite{meng2024simpo}. 

\subsubsection{Normalized Log Probability Contrast}

Normalized log probability contrast is the case where $f\left( {{\pi _\theta }({{\mathbf{y}}}|{{\mathbf{x}}})} \right) = \log \left( {{\pi _\theta }({\mathbf{y}}|{\mathbf{x}})/{\pi _{ref}}({\mathbf{y}}|{\mathbf{x}})} \right)$, the equivalent optimal reward derived from the theoretical form of the optimal RLHF model with KL divergence regularization between policy model ${\pi _\theta }$ and reference model ${\pi _{ref}}$ \cite{dpo}. 
When $D_{con}$ is a pairwise preference dataset with only one negative output $\mathbf{y}^{j,-}$ per sample, $\beta_1=\beta_2=\beta$, and $g(u)=\log\sigma(u)$ for a scalar $u$, the preference-guided contrastive learning objective becomes direct policy optimization (DPO) \cite{dpo}. Derived from the reward modeling objective with equivalent optimal reward, DPO can align LLMs with preference signals directly from human feedback. \citet{rafailov2024r} further derive token-level DPO, enabling the application of token-wise preferences. 
\citet{chen2024selfplay} iterate DPO optimization, using the optimized model in the last iteration as the reference model in the current iteration.
Based on DPO, f-DPO \cite{wang2023beyond} extends the KL divergence constraint in the vanilla RLHF objective to a general form. 
$\Psi$PO \cite{azar2023general} instead extends the reward optimization objective in RLHF, forming another generalized version of DPO. 
Other extensions to DPO include applying the optimal reward to derive contrastive learning objectives for listwise ranking data \cite{liu2024lipo} and preference data with numerical reward scores \cite{mao2024dont,ji2024efficient}.
KTO \cite{ethayarajh2024kto} and ULMA \cite{ulma} further derive pointwise normalized log probability optimization approaches for single positive or negative data samples. KTO introduces a reference point of the equivalent optimal rewards for contrast, while ULMA selects zero as the reference point. 
Besides, \citet{liu2023statistical} demonstrate the effectiveness of normalized log probability contrast, outperforming raw log probability contrast in dialogue experiments.
For the selection of the reference model, ORPO \cite{hong2024orpo} directly takes ${\pi _{ref}} ({\mathbf{y}}|{\mathbf{x}}) = 1 - {\pi _\theta } ({\mathbf{y}}|{\mathbf{x}})$ instead of utilizing an extra fixed reference model other than the policy model, formulating the optimal reward as a log odds function. The log odds function also provides strong adaptation for positives and gentle penalties for negatives, enabling ORPO to eliminate the need for SFT. 

\subsection{Preference-conditioned Fine-tuning and Generation}
Beyond methods using numerical human preference signals, an alternative paradigm is to utilize text tokens, especially natural language texts to condition auto-regressive LLMs toward alignment with human intentions, as shown in the bottom right of Fig. \ref{fig:util}. The formalization of preference-conditioned fine-tuning is shown as follows:
\begin{equation}
    \mathop {\max }\limits_\theta  \sum\limits_{j = 1}^{|{D_{con}}|} {\log {\pi _\theta }({{\mathbf{y}}^j}|{{\mathbf{x}}^j},{{\mathbf{c}}^j})},
\end{equation}
where ${\mathbf{c}}^j$ is the preference conditioning texts. According to the positive or negative label of ${\mathbf{y}}^j$, ${\mathbf{c}}^j$ is differently formatted in ${\mathbf{c}}^{j,+}$ or ${\mathbf{c}}^{j,-}$, respectively. For the preference-conditioned generation, the preference condition is always positive, as formalized below:
\begin{equation}
    {\mathbf{y}} \sim {\pi _\theta }({\mathbf{y}}|{\mathbf{x}},{{\mathbf{c}}^ + }).
\end{equation}
As an example, OpenChat \cite{openchat} uses different dialogue role prefixes to differentiate training data of varying quality, combined with the SFT objective weighted by varying predefined reward scores. In generation, the dialogue role prefix is fixed to the high-quality one. 
\citet{xu2023reasons} prepend the error judgments to the negative outputs for SFT, while positives are retained as-is. 
\citet{woh} reformat the inputs to fit the preference judgments of outputs from a reward model. 
\citet{liu2024chain} chain positives and negatives from each sample prepended with corresponding preference conditioning texts for SFT, while generations are sampled with positive conditioning texts.
Besides natural language texts, \citet{hu2023aligning} directly prepend the reward scores in text form to the training samples for SFT while setting the reward score prefix to the possible maximum for generation. 
\citet{steerlm} further incorporate multiple attribute scores in text form for fine-tuning, while the generation prefix is sampled from recorded combinations of attribute scores in which one of the scores is full. 

\section{Evaluation}
\label{sec:eval}

\begin{table*}
\centering
\caption{Summary of LLM evaluation benchmarks}
\begin{tabular}{c|c|c|c}
\toprule
\thead{Benchmark} & \thead{Evaluation Approach} & \thead{Annotation} & \thead{Task} \\
\midrule
Koala\footnotemark[2] & Open-formed & Unlabeled & Instruction Following \\
Vicuna-bench\footnotemark[3] & Open-formed & Unlabeled & Instruction Following \\
HH-RLHF \cite{anth-llm} & Open-formed & Reference responses & Instruction Following \\
\textsc{Self-Instruct} \cite{wang2022self} & Open-Formed & Reference responses & Instruction Following \\
AlpacaFarm \cite{farm} & Open-formed & Reference responses & Instruction Following \\
MT-bench \cite{zheng2023judging} & Open-formed & Reference responses & Instruction Following \\
\midrule
MMLU \cite{hendrycks2020measuring} & Automatic Evaluations & Golden answers & General Ability Tests, Exams \\
AGIEval \cite{zhong2023agieval} & Automatic Evaluations & Golden answers & General Ability Tests, Exams \\
\textsc{Sup-NatInst} \cite{superni} & Automatic Evaluations & Golden answers & General Ability Tests, NLP tasks \\
Big-Bench-Hard \cite{suzgun2022challenging} & Automatic Evaluations & Golden answers & General Ability Tests, NLP tasks \\
TruthfulQA \cite{truthqa} & Automatic Evaluations & Golden answers & Specific Ability Test, Truthfulness \\
HumanEval \cite{humanev} & Automatic Evaluations & Golden answers & Specific Ability Test, Coding \\
GSM8K \cite{gsm8k} & Automatic Evaluations & Golden answers & Specific Ability Test, Math Reasoning \\
IMDb \cite{imdb} & Automatic Evaluations & Golden answers & Downstream Task, Sentiment Control \\
Reddit TL;DR \cite{tldr} & Automatic Evaluations & Golden answers & Downstream Task, Summarization \\
CNN/DailyMail \cite{cnndm} & Automatic Evaluations & Golden answers & Downstream Task, Summarization \\
\bottomrule
\end{tabular}
\label{tab:eval}
\end{table*}
\footnotetext[2]{\url{https://github.com/arnav-gudibande/koala-test-set/}}
\footnotetext[3]{\url{https://github.com/lm-sys/vicuna-blog-eval/}}

In this section, we summarize the evaluation approaches and benchmarks commonly used to assess the alignment with the human intentions of LLMs. The prevailing approaches can be categorized into three major types: open-form benchmarks, automatic evaluations, and qualitative analyses. 
Commonly used benchmarks are shown in Table \ref{tab:eval}. 

\subsection{Open-form Benchmarks}

Open-form benchmarks are evaluation datasets containing unlabeled instructions or instructions with responses generated from a reference LLM. These human-collected benchmarks focus on assessing the instruction-following ability of an LLM, directly reflecting its alignment with human intentions. As no golden answers can be annotated for these open-domain generation tasks, most automatic metrics based on similarity or accuracy are not applicable. The prevailing evaluation approaches on open-form benchmarks include human evaluation, LLM-based evaluation, and reward model evaluation.

\subsubsection{Human evaluation}

Human evaluation directly reflects human preferences on LLM-generated responses. Most human evaluations are conducted in the form of win-lose comparisons and Likert scales. Although important in LLM evaluation, its downsides, as previously discussed in Subsection \ref{subsec:fbfmod}, are the high costs and the long time it takes. Besides, it is also hard to reproduce the evaluation results.

\subsubsection{LLM-based evaluation}

LLM-based evaluation can be used to evaluate open-form benchmarks at scale with lower costs, regarding the powerful commercial LLMs such as GPT-4 \cite{gpt4} as simulations of humans. Some recent works also attempt to develop LLMs dedicated to evaluations \cite{pandalm,prometheus,li2024generative}. Still, it is subjected to the inherent issues of LLMs such as positional bias \cite{wang2023large}.

\subsubsection{Reward Model Evaluation}

Reward models can serve as a ``surrogate'' of human preferences, as discussed in Subsection \ref{subsec:rewmod}. Therefore, evaluation with reward models is also feasible. 
For example, \citet{ramamurthy2023is} use dedicated reward models to evaluate LLM performance on specific generative tasks. \cite{rrhf} also incorporate a pre-trained reward model for LLM evaluation on open-form benchmarks.
On the other hand, the reliability of reward model evaluation remains unclear, especially for out-of-distribution evaluations and reward hacking responses.

\subsection{Automatic Evaluations}

Automatic evaluations are mostly conducted through labeled benchmark datasets, evaluating how LLMs are aligned with automatic metrics. These evaluations focus on the natural language understanding abilities of LLMs, which can be further divided into general ability tests, specific ability tests, and downstream task benchmarks. 

\subsubsection{General ability tests}

General ability tests aim to evaluate understanding abilities and the overall knowledge of LLMs. They are usually collections of questions appearing in various exams of humans \cite{hendrycks2020measuring,zhong2023agieval} or diverse sets of NLP tasks \cite{superni,suzgun2022challenging}.

\subsubsection{Specific ability tests}

Specific ability tests are dedicated to evaluating the performance of LLMs in specific aspects, such as truthfulness \cite{truthqa}, coding \cite{humanev}, and math reasoning \cite{gsm8k}.

\subsubsection{Downstream tasks}

Downstream task benchmarks evaluate the performance of LLMs on specific generative downstream tasks, such as summarization \cite{tldr,cnndm}. These evaluations are usually conducted for rapid preliminary validations of the performance of LLMs.

\subsection{Qualitative Analyses}

Qualitative analyses complement the above quantitative assessment, illustrating the effects of LLM alignment fine-tuning as well as the potential issues.
The showcase examples should be randomly selected from the responses of the LLMs, thereby avoiding any potential bias caused by cherry-picking.

\section{Conclusion and Outlooks}
\label{sec:conc}

In this survey, we present a comprehensive review of the development timeline and recent advances in human preference learning for LLMs, covering aspects including feedback sources, feedback formats, preference modeling, preference usage, and evaluation. From this survey, we can conclude that: i) {\em simulating humans with LLMs} is feasible and desirable for both human preference collection and human intention alignment evaluation; ii) while {\em numerical preference models} are still the mainstream, human preference modeling in other forms are also inspiring; iii) as typical approaches for RL-based preference usage, especially PPO, are powerful but unstable, {\em improvements or alternatives to RL methods} are both studied. Based on the reviewed progress, we finally discuss several existing challenges and promising areas in human preference learning to align LLMs with human intentions.
\begin{itemize}

    \item {\em Pluralistic human preference learning:} Although humans mostly agree with some general principles of preference, different groups of humans may have varied preferences. \citet{sorensen2024roadmap} discuss possible ways to align AI systems with pluralistic human preferences. \cite{bakker2022ftlmagree} train an LLM aligning to the consensus of most people from diverse human preferences. \citet{gpo} train an LLM with modified Transformer architecture that can generalize to the preferences of different groups with in-context examples. However, it is still worth studying how to simultaneously model general human preferences and preferences of different groups with zero-shot generalizability, satisfying diverse and subtle preferences inferred from human queries.
    
    \item {\em Scalable oversight for aligning LLMs:} Preference learning for LLMs has improved the abilities of LLMs to human levels with the guidance of human preferences. However, when AI systems such as LLMs are more capable than humans, new approaches that scale up the ability of human supervision are required to align them with human intentions, namely scalable oversight \cite{amodei2016concrete}. \citet{bowman2022measuring} list a few potential approaches to scalable oversight, including plain model interaction, debate \cite{irving2018ai}, amplification \cite{christiano2018supervising}, recursive reward modeling \cite{leike2018scalable}, etc., together with a preliminary experiment that validates human-model interaction for scalable oversight. \cite{burns2023weaktostrong} test the weak-to-strong generalization ability by learning a stronger LLM supervised by a weaker language model, indicating the possibility of scalable oversight over superhuman LLMs. More empirical research is required in this promising direction.
    
    \item {\em Language-agnostic LLM alignment:} Intuitively, a range of abilities connected with human intelligence should be language-agnostic, such as reasoning \cite{she2024mapo}. However, existing studies \cite{chen2023breaking} demonstrated that current LLMs are mostly more capable of reasoning in certain languages such as English. \citet{muennighoff-etal-2023-crosslingual} attempt to elicit language-agnostic ability learning for LLMs by SFT on multilingual instruction dataset. \citet{she2024mapo} mitigate the performance difference of languages in math reasoning with preference learning approaches by utilizing the consistency between answers in different languages with the help of an off-the-shelf translation model as preference. Further research can be conducted on mitigating or eliminating the overall capability gap in languages for LLMs by language-agnostic LLM alignment.
    
    \item {\em Alignment with multi-modal complement:} Large multi-modal models such as GPT-4V \footnote[4]{\url{https://openai.com/research/gpt-4v-system-card}} extend LLMs with the ability to perceive and understand multi-modal information for more downstream applications beyond language. Therefore, aligning large multi-modal models with human intention is also necessary and crucial. 
    Recent large multi-modal model alignment approaches usually align features between modalities and model behaviors with human intentions separately, without explicit reliance on the complement relation between multi-modal input and instructions with intentions \cite{vit,instblip,jiang2024hallucination}. \citet{sun2023aligning} utilize image captions to condition the reward model for RLHF, and more in-depth research in utilizing the complement relation between modalities to align modality features and model behaviors simultaneously can be conducted.
    
    \item {\em Comprehensive assessment of LLM alignment progress:} Among the LLM evaluation approaches discussed in \ref{sec:eval}, instruction following benchmarks and general ability tests are relatively more comprehensive.  
    However, neither can comprehensively assess LLM alignment progress. 
    General ability tests are mostly formed in multiple-choice questions, efficient to evaluate at scale but unable to evaluate the generative abilities of LLMs. 
    Instruction following benchmarks require humans or powerful commercial LLMs such as GPT-4 for relatively direct evaluation of human intentions. Therefore, the scale of instructions for evaluation is usually limited, considering efficiency and cost. 
    New evaluation benchmarks and approaches combining the scale and efficiency of general ability tests and direct evaluation of human intentions are favorable for comprehensively revealing the progress when aligning LLMs.
    
    \item {\em Empirically researching deceptive alignment:} Deceptive alignment is where AI systems game the training signals to make them look aligned while optimizing for goals not intended by humans.\footnote[5]{\url{https://www.alignmentforum.org/posts/zthDPAjh9w6Ytbeks/deceptive-alignment}} 
    Currently, little empirical research is conducted on this topic, while its concern is growing as its severe consequences are widely acknowledged.\footnote[6]{\url{https://www.lesswrong.com/posts/Km9sHjHTsBdbgwKyi/monitoring-for-deceptive-alignment}} 
    Validation experiments validating its existence and alignment approaches may be substantial progress in this topic.
\end{itemize}
 
\bibliographystyle{IEEEtran}
\bibliography{IEEEabrv,tpami}

\vfill

\end{document}